\documentclass[10pt,twocolumn,letterpaper]{article}

\usepackage{cvpr}              %
\newcommand{\myparagraph}[1]{\vspace{0.1em}\noindent\textbf{#1}}

\usepackage[dvipsnames]{xcolor}

\definecolor{cvprblue}{rgb}{0.21,0.49,0.74}
\usepackage[pagebackref,breaklinks,colorlinks,citecolor=cvprblue]{hyperref}

\def\paperID{949} %
\def\confName{CVPR}
\def\confYear{2024}

\title{HiFi4G: High-Fidelity Human Performance Rendering via Compact Gaussian Splatting}
\author{Yuheng Jiang$^{1,2}$ \;\, Zhehao Shen$^{1}$ \;\, Penghao Wang$^{1}$ \;\, Zhuo Su$^{3}$ \;\, Yu Hong$^{1}$ \;\, \\ Yingliang Zhang$^{4}$ \;\, Jingyi Yu$^{1}$ \;\, Lan Xu$^{1}$} 

\makeatletter
\let\@oldmaketitle\@maketitle%
\renewcommand{\@maketitle}{
	\@oldmaketitle%
	\centering
	\vspace{-8mm}
	{\large \textsuperscript{1}ShanghaiTech University}\quad \quad
	{\large \textsuperscript{2}NeuDim}\quad \quad
	{\large \textsuperscript{3}ByteDance}\quad \quad
	{\large \textsuperscript{4}DGene}\quad \quad

	\vspace{8mm}
}

\usepackage{graphicx}
\usepackage{amsmath}
\usepackage{multirow}
\usepackage{booktabs}
\usepackage{lipsum}
\usepackage{colortbl}

\newcommand{\bestCellColor}[1]{\cellcolor[rgb]{.866,.945, 0.831}#1}

\newcommand{\secondBestCellColor}[1]{\cellcolor[rgb]{1, 0.98, 0.83}#1}

\begin{document}

\twocolumn[{
  \renewcommand\twocolumn[1][]{#1}
  \maketitle
  \begin{center}
  \vspace{-2ex}
  \includegraphics[width=0.99\textwidth]{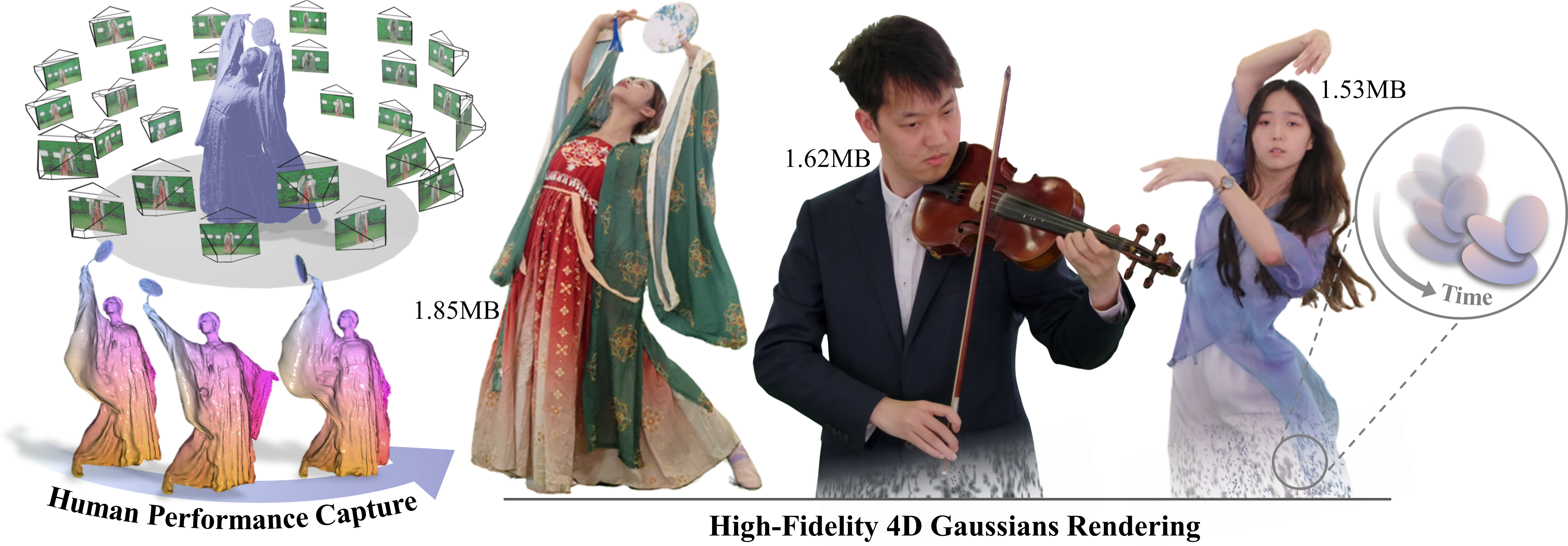}
  \vspace{-2ex}
  \captionof{figure}{\small{\textbf{High-fidelity rendering with our compact Gaussian Splatting}.} From multi-view human performance video, HiFi4G marries the traditional non-rigid fusion with differentiable rasterization advance to efficiently produce compact 4D assets.}
  \label{fig:teaser}
  \end{center}
}]

\begin{abstract}
  
We have recently seen tremendous progress in photo-real human modeling and rendering. Yet, efficiently rendering realistic human performance and integrating it into the rasterization pipeline remains challenging. 
In this paper, we present HiFi4G, an explicit and compact Gaussian-based approach for high-fidelity human performance rendering from dense footage. Our core intuition is to marry the 3D Gaussian representation with non-rigid tracking, achieving a compact and compression-friendly representation.
We first propose a dual-graph mechanism to obtain motion priors, with a coarse deformation graph for effective initialization and a fine-grained Gaussian graph to enforce subsequent constraints. 
Then, we utilize a 4D Gaussian optimization scheme with adaptive spatial-temporal regularizers to effectively balance the non-rigid prior and Gaussian updating.
We also present a companion compression scheme with residual compensation for immersive experiences on 
various platforms. It achieves a substantial compression rate of approximately 25 times, with less than 2MB of storage per frame.
Extensive experiments demonstrate the effectiveness of our approach, which significantly outperforms existing approaches in terms of optimization speed, rendering quality, and storage overhead. Project page: \href{https://nowheretrix.github.io/HiFi4G/}{https://nowheretrix.github.io/HiFi4G/}.

\end{abstract}

\section{Introduction}
Volumetric recording and realistic rendering of 4D (space-time) human performance diminish the boundaries between viewers and performers. It brings numerous immersive experiences like telepresence or tele-education in VR/AR.%

Early solutions~\cite{collet2015high, dou2016fusion4d, motion2fusion, TotalCapture} reconstruct textured meshes from captured videos by explicitly leveraging non-rigid registration~\cite{sumner2007embedded,newcombe2015dynamicfusion}. Yet, they remain vulnerable to occlusions and lack of textures which cause holes and noise in the reconstruction results. Recent neural advances, represented by NeRF~\cite{nerf}, bypass explicit reconstruction and instead optimize a coordinate-based multi-layer perceptron (MLP) to conduct volume rendering at photo-realism. Some dynamic variants~\cite{tretschk2021nonrigid, park2021nerfies, park2021hypernerf, ARAH:ECCV:2022, Gafni_2021_CVPR, peng2021animatable} of NeRF attempt to maintain a canonical feature space to reproduce features in each live frame with an extra implicit deformation field. However, such a canonical design is fragile to large motions or topology changes. Recent approaches~\cite{wang2023neus2,song2023nerfplayer,isik2023humanrf} remove the deformation fields and compactly represent the 4D feature grid through planar factorization~\cite{chen2022tensorf,fridovich2023k} or Hash-encoding~\cite{muller2022instant}. They significantly accelerate both the training and rendering speed for interactive applications but the challenges of runtime memory and storage still exist.
The recent 3D Gaussian Splatting (3DGS)~\cite{kerbl20233d} strikes back to an explicit paradigm for static scene representation. Based on GPU-friendly rasterization of 3D Gaussian primitives, it allows real-time and high-quality radiance field rendering unseen before. Various concurrent works~\cite{luiten2023dynamic,wu20234d,yang2023deformable, yang2023real}  adapt 3DGS for dynamic scenes. Some~\cite{luiten2023dynamic} focus on extracting the non-rigid motions from dynamic Gaussians yet sacrificing the rendering quality. Others~\cite{wu20234d,yang2023deformable} adopt extra implicit deformation fields to compensate for the motion information, and hence fall short of handling long-duration motions and lose the explicit and GPU-friendly beauty of the original 3DGS.

In this paper, we present \textit{HiFi4G} -- a totally explicit and compact Gaussian-based approach for high-fidelity 4D human performance rendering from dense footage (see Fig.~\ref{fig:teaser}). Our key idea is to marry the 3D Gaussian representation~\cite{kerbl20233d} with non-rigid tracking, so as to explicitly disentangle motion and appearance information for a compact and compression-friendly representation.
HiFi4G significantly outperforms existing implicit rendering approaches, in terms of optimization speed, rendering quality, and storage overhead. Our explicit representation also enables seamlessly integrating our results into the GPU-based rasterization pipeline.%
, i.e., immersively watching high-fidelity human performances with VR headsets.

To organically bridge Gaussian representation with non-rigid tracking, we first introduce a dual-graph mechanism, which consists of a coarse deformation graph and a fine-grained Gaussian graph. For the former, we obtain per-frame geometry proxy via the NeuS2~\cite{wang2023neus2} and then employ embedded deformation (ED)~\cite{sumner2007embedded} in a key-frame manner. Such an explicit tracking process splits the sequence into segments and provides rich motion prior within each segment. Analogous to the key-volume update~\cite{dou2016fusion4d}, we follow 3DGS to prune the incorrect Gaussians from the previous segment and update new ones to restrict the number of Gaussians in the current segment. 
Then, we build a fine-grained Gaussian graph and interpolate the motion of each Gaussian from the coarse ED graph for subsequent initialization. Na\"{i}vely warping the Gaussian graph with the ED graph and splatting it onto screen space will cause severe unnatural artifacts, while continuous optimization without any constraints leads to jittery artifacts.
Thus, we propose a 4D Gaussian optimization scheme to carefully balance the non-rigid motion prior and the updating of Gaussian attributes. We adopt a temporal regularizer to enforce the appearance attributes of each Gaussian, i.e., spherical harmonic (SH), opacity, and scaling coefficients, to be consistent. We also propose a smooth term for the motion attributes (position and rotation) to produce locally as-rigid-as-possible motions between the adjacent Gaussians. These regularizers are further enhanced with an adaptive weighting mechanism to penalize the flicking artifacts on the regions with slight non-rigid motions. 
Once optimized, we obtain spatial-temporally compact 4D Gaussians. To make our HiFi4G practical for users, we demonstrate a companion compression scheme that follows standard residual compensation, quantization, and entropy encoding for the Gaussian parameters. 
It achieves a substantial compression rate of approximately 25 times and only requires less than 2 MB storage per frame, enabling immersively viewing human performances on various platforms like VR headsets.

To summarize, our main contributions include:
\begin{itemize} 
	\setlength\itemsep{0em}
	
	\item We present a compact 4D Gaussian representation for human performance rendering, which bridges Gaussian Splatting and non-rigid tracking.

	\item We propose a dual-graph mechanism with various regularization designs to effectively recover spatial-temporally consistent 4D Gaussians. 
	
	\item We showcase a companion compression scheme, supporting immersive experience of human performance with low storage, even under various platforms. 
  	
\end{itemize}

\section{Related Work} 
\noindent{\textbf{Human Performance Capture.}} 
Recently, human performance capture~\cite{jiang2022hifecap, Wang2021CVPR, Prokudin_2023_ICCV, SkiRT:3DV:2022, li20214dcomplete, burov2021dsfn, palafox2021npms, bozic2021neural, Challencap2021, chen2021snarf} has been widely investigated to achieve detailed registration for various applications.
Zollh{\"o}fer~\etal~\cite{zollhofer2014real} capture the rigid template first but DynamicFusion~\cite{newcombe2015dynamicfusion} removes this explicit template prior and enables real-time performance which benefits from the GPU solvers. Guo~\etal~\cite{guo2015robust} model the geometry, surface albedo, and appearance on the reference volume, generating impressive tracking results. 
Fusion4d~\cite{dou2016fusion4d} and Motion2fusion~\cite{motion2fusion} rely on a key-frame-based strategy to handle topological changes.
Based on the human parametric model~\cite{SMPL2015}, DoubleFusion~\cite{DoubleFusion} proposes a two-layer representation for more robust scene capture, while Xu~\etal~\cite{UnstructureLan} extend it to sparse view setup. Su~\etal~\cite{robustfusion, su2022robustfusionPlus} further address the challenging motions and human-object interaction scenarios. 
Additionally, several studies~\cite{yu2021function4d, jiang2022neuralhofusion, li2021posefusion,li2022avatarcap} combine explicit volumetric fusion and implicit modeling to capture more dynamic details. Nevertheless, these methods primarily focus on detailed geometry rather than high-quality texture. Comparably, our approach bridges volumetric capture and recent differentiable rasterization advances, achieving high-fidelity human performance rendering.

\noindent{\textbf{Dynamic Scene Modeling.}} In the domain of dynamic scene representation, various approaches~\cite{liu2020NeuralHumanRendering, zheng2023avatarrex, lin2022efficient, zhang2023closet, suo2021neuralhumanfvv, sun2021HOI-FVV, luo2022artemis} have been proposed to address this challenge. 
D-NeRF~\cite{pumarola2020d} and Non-rigid NeRF~\cite{tretschk2021nonrigid} utilize a displacement field to represent the motion, while Neuralbody~\cite{peng2021neural} uses latent codes anchored to SMPL~\cite{SMPL2015} vertices. Humannerfs ~\cite{zhao2022humannerf, weng_humannerf_2022_cvpr} combine the SMPL with a deformation net. TAVA~\cite{li2022tava} and X-avatar~\cite{shen2023x} learn the skinning weight through root-finding. NDR~\cite{Cai2022NDR} defines a bijective function that satisfies the cycle consistency. With recent advancements in Instant-NGP~\cite{muller2022instant}, some works~\cite{wang2023neus2, song2023nerfplayer, jiang2023instant, isik2023humanrf, lin2022efficient} demonstrate the efficient training and rendering speed. However, most methods produce blurriness, particularly in high-frequency regions. 
The recent 3DGS~\cite{kerbl20233d} strikes back to an explicit paradigm for high-performance static scene representation. However, per-frame 3DGS disregards temporal consistency, resulting in visual jitteriness. Some concurrent studies~\cite{luiten2023dynamic, wu20234d,yang2023deformable, yang2023real} adapt 3DGS for dynamic scenes. Yet, these methods typically offer real-time performance only at low resolution and are not equipped to handle large motions. In contrast, HiFi4G leverages dual-graph to generate compact 4D Gaussians, enabling high-fidelity real-time rendering with challenging motions.

\noindent{\textbf{Compact Representation.}}
Compact representation plays a pivotal role in dynamic rendering, engaging the interest of numerous researchers. 
A series of works are proposed for early point cloud compression with Octree~\cite{schnabel2006octree, thanou2016graph},  Wavelet~\cite{nadenau2003wavelet}. These are formalized into MPEG-PCC~\cite{schwarz2018emerging} standards by the Moving Picture Experts Group(MPEG), which are categorized into video-based(VPCC) and geometry-based(GPCC).
Following, learning-based methods~\cite{quach2019learning, quach2020improved, liang2022transpcc} emerge, focusing on enhancing efficiency. 
For neural fields, several studies introduce compact neural representations through tensor~\cite{chen2022tensorf} and scene~\cite{song2023nerfplayer} decomposition, tri-planes~\cite{reiser2023merf, hu2023tri} and multi-planes~\cite{shao2023tensor4d, cao2023hexplane, fridovich2023k}.
Instant-NSR~\cite{zhao2022human} leverages the tracked mesh and texture video while HumanRF~\cite{isik2023humanrf} employs temporal matrix-vector decomposition. 
Despite their advancements, these methods often compromise rendering quality and speed to minimize storage requirements. Comparably, HiFi4G achieves a substantial compression rate of approximately 25 times and only requires less than 2 MB storage per frame to enable high-quality rendering results.

    \begin{figure*}[t] 
	\begin{center} 
		\includegraphics[width=\linewidth]{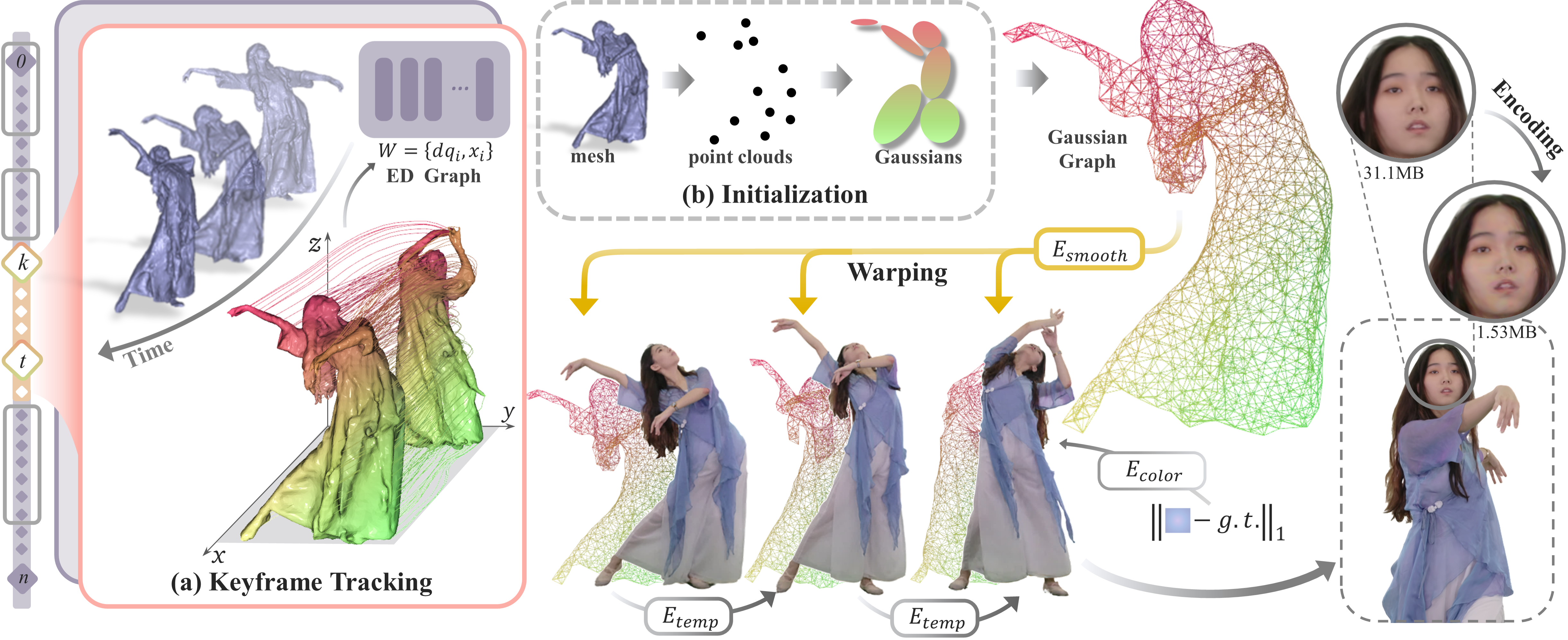} 
	\end{center} 
    \vspace{-20pt}  
	\caption{\noindent{\bf Overview of HiFi4G.}  (a) The non-rigid tracking establishes a \noindent{\bf coarse deformation graph} and tracks the motions for Gaussian optimization. (b) HiFi4G initializes first frame Gaussians from NeuS2 and constructs a \noindent{\bf fine-grained Gaussian graph} to enhance temporal coherence. We then employ the ED graph to warp 4D Gaussians, applying both $E_{\mathrm{smooth}}$ and $E_{\mathrm{temp}}$ constraints to the Gaussian graph, which yields spatial-temporally compact and compression-friendly 4D Gaussians, thus facilitating efficient compression.}
	\label{fig:fig_2_overview} 
	\vspace{-8pt}
\end{figure*}

\section{Method}\label{sec:algorithm}

Given human performance videos captured by multi-view pre-calibrated and synchronized RGB cameras, HiFi4G integrates recent advancements in differentiable rasterization with traditional non-rigid tracking,
significantly outperforming existing rendering approaches~\cite{isik2023humanrf, lin2023im4d, zhao2022human, wang2023neus2} in terms of optimization speed, rendering quality, and storage overhead.
The methodology is visually summarized in Fig.~\ref{fig:fig_2_overview}.
Our approach starts with a dual graph mechanism, which consists of a coarse deformation graph and a fine-grained Gaussian graph, detailed in Sec~\ref{sec:init}. 
Subsequently, this representation is employed along with corresponding temporal and smooth regularization, leading to the generation of spatial-temporally compact 4D Gaussians in Sec~\ref{sec:opt}. 
In addition, we introduce a companion compression scheme in Sec~\ref{sec:compression}. This allows for immersive viewing of high-fidelity human performances with a storage requirement of less than 2 MB per frame.

\subsection{Dual Graph Mechanism} \label{sec:init}
We employ a dual graph structure to explicitly disentangle motion and appearance, resulting in a compact and compression-friendly representation. This design facilitates expedited convergence and enhances visual quality.

\noindent{\bf Coarse Deformation Graph.} 
Instead of using an additional implicit deformation network~\cite{wu20234d, yang2023deformable} to handle non-rigid motion, which could potentially affect the high performance and GPU-friendliness of the original 3DGS, we opt for the Embedded Deformation~\cite{sumner2007embedded} to establish model-to-model correspondences by leveraging conventional non-rigid deformation techniques~\cite{newcombe2015dynamicfusion, FlyFusion, guo2017real}.
To achieve this, we first generate per-frame geometry proxies using NeuS2~\cite{wang2023neus2}. We then apply non-rigid tracking to the resulting mesh sequences following a key-frame manner.
Specifically, we parameterize the dynamic motions as an ED graph $W=\{dq_i,x_i\}$, where $x_i$ represents the coordinates of sampled ED nodes in key space, and $dq_i$ denotes the dual quaternions representing the corresponding rigid transformation in $SE(3)$ space. 
Subsequently, we acquire each point $v_c$ using Dual-Quaternion Blending:
\begin{equation} \label{dual-quaternion blending}
{DQB}(v_c) = \sum_{i \in \mathcal{N}(v_c)} w(x_i,v_c)dq_i,
\end{equation}
where $\mathcal{N}(v_c)$ is a set of neighboring ED nodes of $v_c$. and $w(x_i,v_c)$ denotes the influence weight of the $i-th$ node $x_i$ on $v_c$.
At frame $t$, we identify correspondence points between the warped key mesh and the current mesh. Subsequently, we optimize the motion by constructing the terms:
\begin{equation} \label{ed term}
E = \lambda_{\mathrm{data}}E_{\mathrm{data
}} + \lambda_{\mathrm{reg}}E_{\mathrm{reg}},
\end{equation}
Where $E_{\mathrm{data}}$ and $ E_{\mathrm{reg}}$ represent the energies associated with the data term and the regularization term, respectively.
Please refer to ~\cite{newcombe2015dynamicfusion, DoubleFusion} for more details. 
To explicitly handle coarse topological changes and reduce severe misalignment issues, we implement the key-volume strategy as described in~\cite{dou2016fusion4d, FlyFusion}. This strategy involves segmenting the sequence into multiple key volumes. 

\noindent{\bf Fine-grained Gaussian Graph.} 
To bypass the tedious process of creating 3D Gaussians from Structure-from-Motion (SfM) points for each frame, we utilize a more efficient initialization method. For the first frame, 
we construct the 3D Gaussians from the NeuS2 mesh using an importance sampling strategy. We increase the sampling density in the hand and face regions to significantly improve visual quality.
For subsequent keyframes, analogous to the key-volume update strategy, we follow 3DGS to prune incorrect Gaussians from the previous keyframe and densify new ones at the current keyframe. We then restrict the number of Gaussians within the current segment.
Afterward, we establish a fine-grained Gaussian graph, consisting of refined Gaussian kernels for subsequent constraints, determined by the k-nearest neighbors (KNN,  k = 16). In addition, for each Gaussian kernel in the fine-grained graph, we also find the KNN (k = 8) from the ED nodes, which assists in calculating the influence weight for motion interpolation. 
The initialization is still crucial for non-key frames to prevent falling into local optima during the back-propagation of differentiable rasterization.
To this end, we warp the Gaussian graph from the keyframe to other frames within the segment according to the ED nodes' motion interpolation: 
\begin{equation}  \label{eq:warp} 
\begin{split}
\begin{aligned}
     p_{i,t}' &= SE3( {DQB}  (p_{i,k})) p_{i,k}, \\
     q_{i,t}' &= ROT({DQB}  (p_{i,k})) q_{i,k},
\end{aligned}
\end{split}
\end{equation}
$SE3(\cdot)$ converts dual quaternion back into a transformation matrix, while $ROT(\cdot)$ extracts the rotation component from dual quaternion. $p_{i,k}, q_{i,k}$ denote the position and rotation of the $i$-th Gaussian kernel at keyframe $k$, respectively. $p_{i,t}'$ and $q_{i,t}'$ represent the initial position and rotation at frame $t$. They will be further optimized in the subsequent stage.

\subsection{4D Gaussians Optimization} \label{sec:opt}
Directly warping the fine-grained Gaussian graph with tracking prior and splatting it onto screen space can lead to noticeable and unnatural artifacts.
To mitigate this, we do not use the Gaussians' densification and pruning within the segment. Instead, we impose a constraint on their number and execute sequential optimization.

For the frame $t$, 
we categorize attributes for each 4D Gaussian kernel $i$ into two groups:
1). Appearance-aware parameters, which include spherical harmonic $\mathcal{C}_{i,t}$, opacity $\sigma_{i,t}$, and scaling $s_{i,t}$.
2). Motion-aware parameters, which include position $p_{i,t}$ and rotation $q_{i,t}$.
Leveraging the initialization from the warped Gaussian graph reduces the training time to one-third while still yielding vivid results. However, despite incorporating non-rigid tracking priors, we observe notable temporal jitters in the rendered results.
Concurrent studies~\cite{luiten2023dynamic,wu20234d,yang2023deformable} address this issue by decoupling the deformation field from canonical 3D Gaussians. They employ a consistent set of Gaussians across dynamic sequences, which substantially diminishes view-dependent effects and sacrifices rendering quality.
To mitigate temporal jitters while maintaining rendering quality, we introduce temporal and smooth regularization to delicately balance the dual graph prior and the updating of Gaussian attributes, thereby enforcing 4D consistency.
First, we introduce the temporal regularization term $E_{\mathrm{temp}}$. This term promotes coherent appearances by constraining the 4D Gaussian appearance attributes($\mathcal{C}_{i,t}$, $\sigma_{i,t}$, $s_{i,t}$) to be consistent with the previous frame:
\begin{equation}
\begin{split}
E_{\mathrm{temp}} & = \sum_{a \in \{ \mathcal{C}, \sigma, s \}} w_{i,t}  \lambda_{\mathrm{a}} \| a_{i,t} - a_{i,t-1} \|^2_2, 
\end{split}
\end{equation}
$E_{\mathrm{temp}}$ helps reduce jitteriness. However, it may be not sufficient, especially when motion parameters change significantly, particularly in feature-less areas. Moreover, applying this regularization directly to motion attributes $p$ and $q$ can also result in unnatural artifacts.
To address this issue, we introduce a smooth term targeted at the motion attributes ($p_{i,t}$, $q_{i,t}$) within the fine-grained Gaussian graph. We define this term as follows:
\begin{equation}
\begin{aligned}
E_{\mathrm{smooth}}  = &\sum_{i} \sum_{j \in \mathcal{N}(i)} w_{i,t} \| SO3(q_{i,t} * q_{i,t-1}^{-1})\\
&(p_{j,t-1} - p_{i,t-1}) - (p_{j,t} - p_{i,t})  \|_2^2, 
\end{aligned}
\end{equation}
$SO3(\cdot)$ converts a quaternion into a rotation matrix. Kernel $i$ and $j$ are neighbors on the Gaussian graph. The smooth term produces locally as-rigid-as-possible deformations to constrain the consistent 4D Gaussian motion on the spatial-temporal domain. 
Furthermore, it's observed that the Human Visual System is more sensitive to detail changes in static regions as opposed to dynamic ones~\cite{carandini2005we}. Thus, we incorporate an adaptive weight that takes into account the displacement of positions between adjacent frames:
\begin{equation}
\begin{aligned}
w_{i,t} = \exp(-\alpha \| p'_{i,t} - p'_{i,t-1}\|^2), \\
\end{aligned}
\end{equation}
This adaptive weight indicates the degree of motion change in a corresponding local region. It penalizes the flicking artifacts in regions with slight non-rigid motions and reduces penalties in areas with large movements. This significantly improves the visual quality.
Additionally, we employ the photometric loss during the training process:
\begin{equation}
\begin{aligned}
E_{\text {color }} &=\|\hat{\mathbf{C}}-\mathbf{C}\|_1, 
\end{aligned}
\end{equation}
$\hat{\mathbf{C}}$ is the blended color after rasterization and $\mathbf{C}$ is the ground truth. The complete energy is as follows:
\begin{equation}
\begin{aligned}
E = \lambda_{\mathrm{temp}} E_{\text{temp}} + \lambda_{\mathrm{smooth}} E_{\text{smooth}} + \lambda_{\mathrm{color}} E_{\text{color}}. \\
\end{aligned}
\end{equation}

\subsection{Compact 4D Gaussians} \label{sec:compression}

\begin{figure}[tbp] 
	\centering 
	\includegraphics[width=1\linewidth]{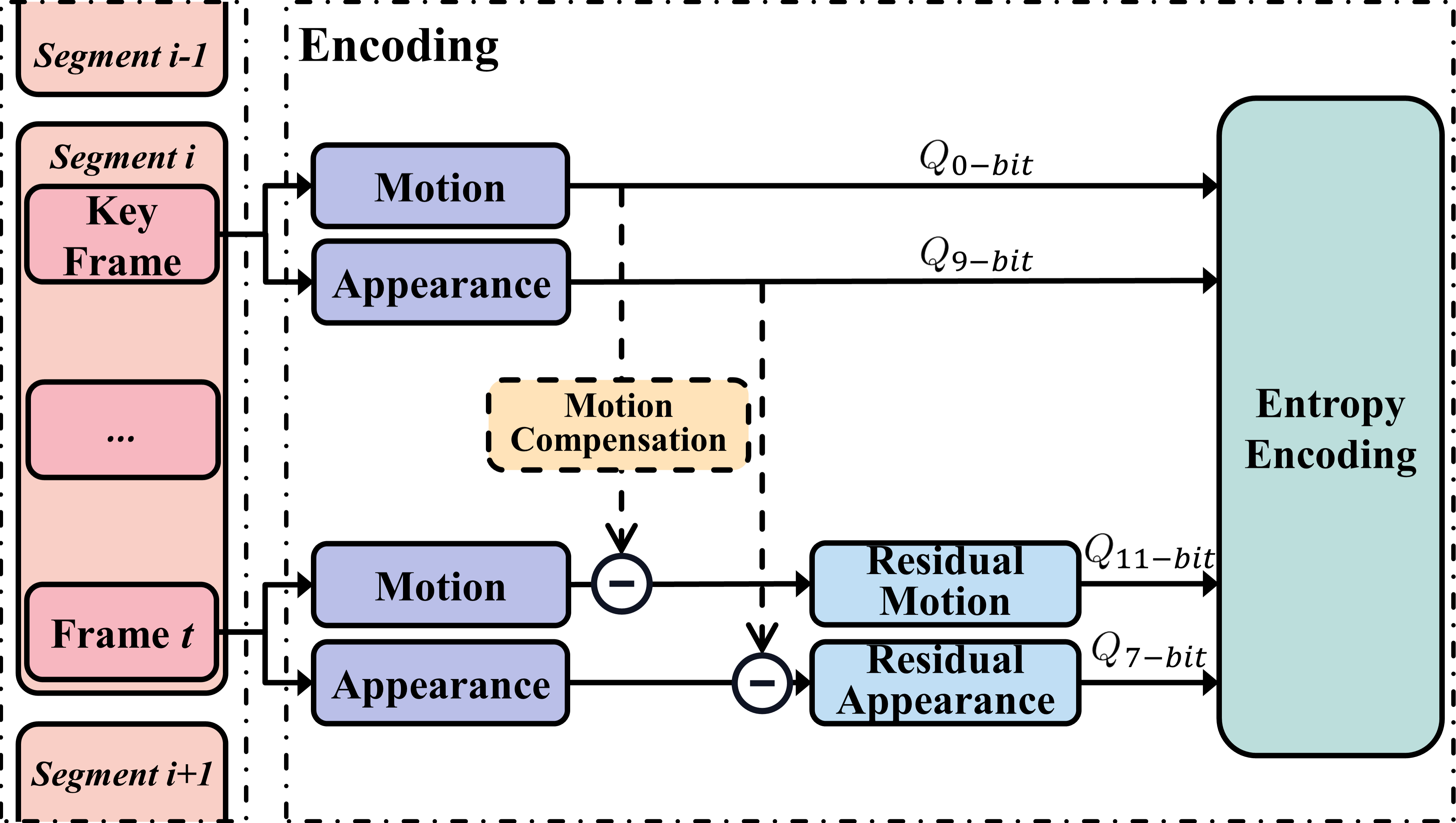} 
	\vspace{-20pt} 
	\caption{Illustration of compression strategy for 4D Gaussians.} 
	\label{fig:compress} 
	\vspace{-15pt} 
\end{figure}

After optimization, we obtain spatial-temporally compact 4D Gaussians, resulting in high-fidelity rendering results. However, each frame requires the same amount of storage as the keyframe. This leads to significant memory consumption and presents challenges when handling lengthy sequences.
To address this problem,  we introduce a companion compression scheme on top of our compact 4D Gaussians. This scheme adheres to the traditional method of residual compensation, quantization, and entropy encoding, as depicted in Fig.~\ref{fig:compress}.

\noindent{\bf Residual Compensation.} 
In contrast to the broad distribution range of the original attributes, we choose to retain the keyframe attributes and calculate residuals for the following frames within the segment. This effectively narrows the range of attributes.
In terms of appearance attributes($\mathcal{C}_{i,t}$, $\sigma_{i,t}$, $s_{i,t}$), the impact of $E_{\text{temp}}$ results in minimal variations. As a result, we can directly derive the residual appearance through subtraction. However, for position $p$ and rotation $q$,  simple subtraction is not sufficient as large motions still exist within a segment. To address this, we employ motion compensation as outlined in Eq.~\ref{dual-quaternion blending} and Eq.~\ref{eq:warp}. We subtract the warped key Gaussians $p_{i,k}', q_{i,k}'$ from $p_{i,t}, q_{i,t}$, ensuring a narrower range.

\noindent{\bf Quantization.}
We scale and round attribute values based on their range and quantization bits $Q_{bit}$, making the data ready for entropy encoding.

\begin{figure*}[htbp] 
	\begin{center} 
		\includegraphics[width=1.0\linewidth]{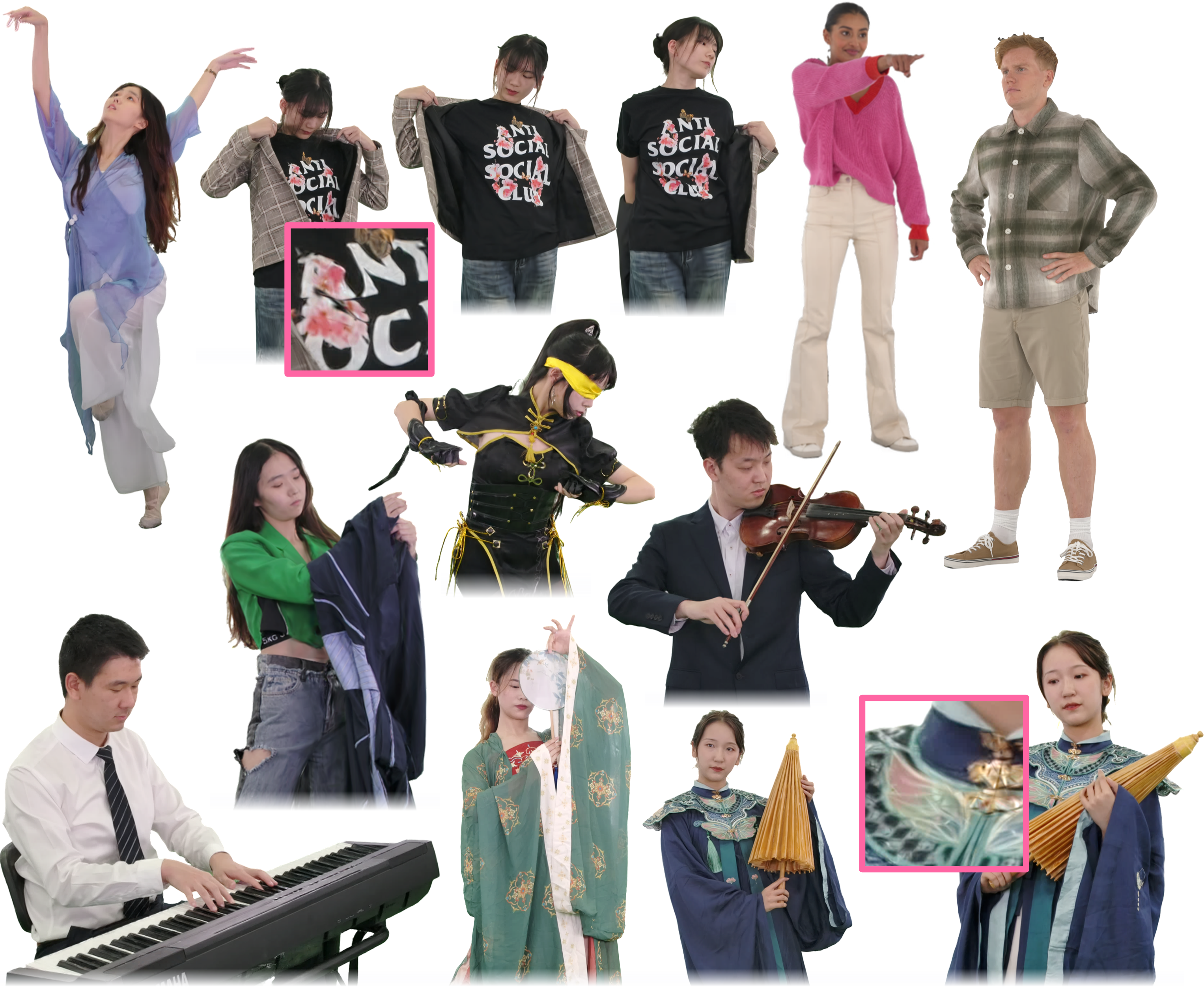} 
	\end{center} 
	\vspace{-20pt}
	\caption{Gallery of our results. HiFi4G delivers real-time high-fidelity rendering of human performance across challenging motions, 
 such as ``playing instruments'', ``dancing'' and ``changing clothes''.
 } 
	\label{fig:gallery}
	\vspace{-10pt}
\end{figure*}
\noindent{\bf Entropy Encoding.}
Residual computation combined with motion compensation yields a residual distribution for attributes that cluster around zero. To leverage this distribution for real-time encoding and decoding, we apply the Ranged Arithmetic Numerical System (RANS)~\cite{duda2013asymmetric}. RANS enhances compression by taking advantage of the distribution's skewness, a key factor for meeting the high-performance demands of HiFi4G.
We compress our data by calculating the frequency of each quantized attribute and constructing a frequency distribution. This distribution helps to encode each attribute efficiently using the RANS algorithm, where each attribute and the current state of the encoder are processed to update the state, representing the encoded data sequence. The final state is stored as an integer stream for subsequent decoding.
This compression scheme achieves a substantial compression rate of approximately 25 times, reducing the storage requirement to less than 2 MB per frame. This capability facilitates the immersive viewing of high-fidelity human performances on various platforms, including VR/AR HMDs.

\section{Implementation Details}\label{sec:detail} 

First, we use the background matting~\cite{lin2021real} to extract the foreground masks from all captured frames. 
For global initialization, we use~\cite{OpenPose} to estimate the hand and face regions for importance sampling. The sampling ratio across the body, hands, and face regions is approximately 8:1:1.
We perform 30000 training iterations with densification and pruning on the keyframes, followed by resetting the tracking and reconstructing the dual-graph. For non-key frames, training iterations are reduced to 9000. In the optimization stage, we use the following empirically determined parameters: $\alpha = 50,  \lambda_{\mathcal{C}} = 1, \lambda_{\sigma} = 0.05,   \lambda_{s}=0.05, \lambda_{\mathrm{smooth}} = 0.002, \lambda_{\mathrm{temp}} = 0.0005, \lambda_{\mathrm{color}} = 1.0 $. 
During compression, we first quantize the appearance attributes, then fix these parameters and fine-tune motion $p$ and $q$ of 4D Gaussians over an additional 1000 iterations. Afterward, we quantize the motion.
We apply different precision levels for various attributes to balance storage and quality. For the keyframes, we keep the motion uncompressed(0-bit) and apply 9-bit quantization for appearance.
For non-key frames, we use 11-bit quantization for motion and 7-bit quantization for appearance due to their more compact range.

\begin{figure*}[htbp] 
	\begin{center} 
		\includegraphics[width=0.99\linewidth]{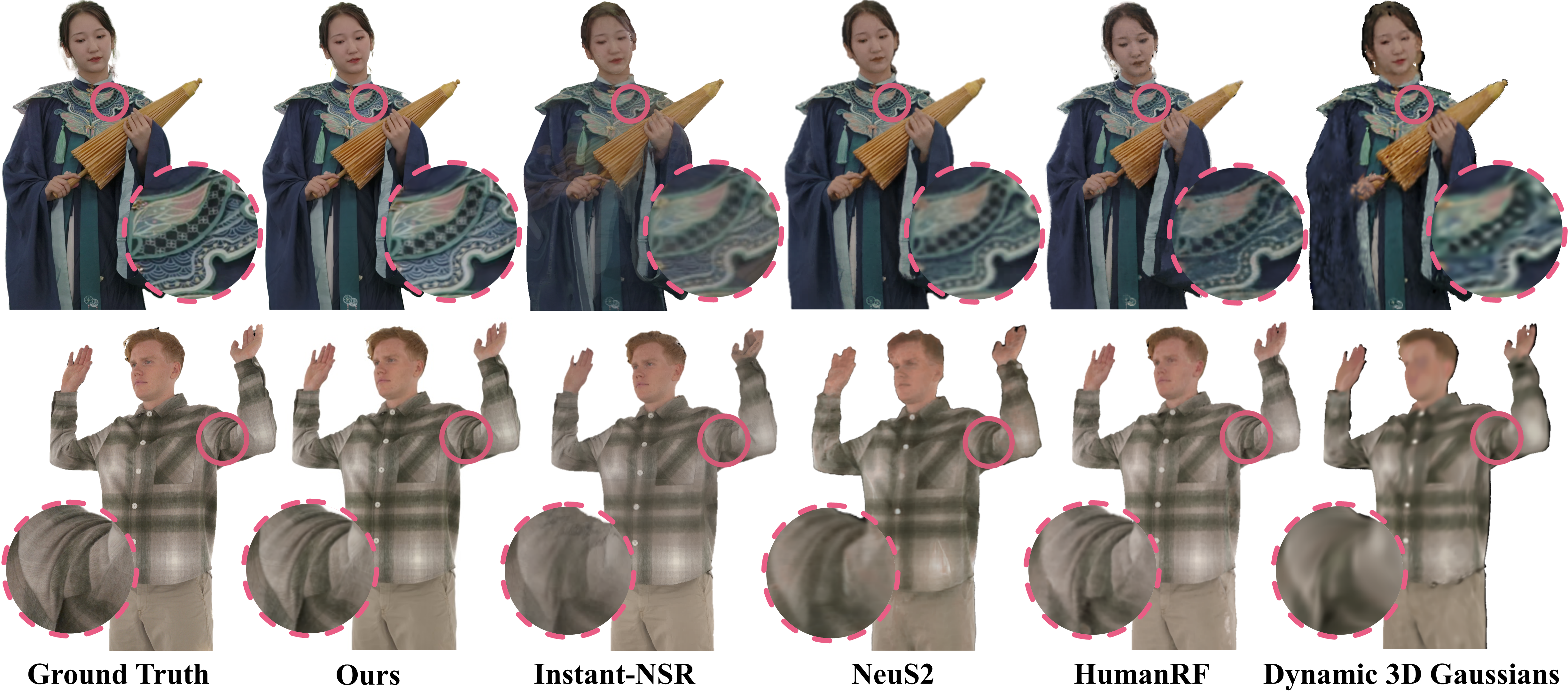} 
	\end{center} 
	\vspace{-16pt}
	\caption{Qualitative comparison of our method against Instant-NSR~\cite{zhao2022human}, NeuS2~\cite{wang2023neus2}, HumanRF~\cite{isik2023humanrf} and Dynamic 3D Gaussians~\cite{luiten2023dynamic} on both our dataset and ActorsHQ~\cite{isik2023humanrf}. Our method achieves the highest rendering quality.}
	\label{fig:fig_comp_1}
\end{figure*}

\section{Experimental Results} 
To demonstrate the capabilities of HiFi4G, we deploy 81 pre-calibrated Z-CAM cinema cameras to capture complex human performances with a resolution of 3840 $\times$ 2160 at 30 fps, and then evaluate our method. 
The dataset covers a variety of costumes, from traditional Chinese attire to casual clothes and cosplay. It also includes a wide range of activities such as dance, fitness, and interaction with various objects.
As shown in Fig.~\ref{fig:gallery}, HiFi4G enables real-time, high-fidelity rendering of human performance in high resolution. It effectively handles complex motions like playing instruments, dancing, and changing clothes. Additionally, our explicit representation allows for seamless integration of our results into the GPU-based rasterization pipeline. We highly recommend watching the supplementary videos.

\subsection{Comparison} 
We compare HiFi4G with the SOTA methods including Instant-NSR~\cite{zhao2022human}, NeuS2~\cite{wang2023neus2}, HumanRF~\cite{isik2023humanrf} and concurrent work Dynamic 3D Gaussians~\cite{luiten2023dynamic} on our captured dataset and ActorsHQ~\cite{isik2023humanrf}.
As depicted in Fig.~\ref{fig:fig_comp_1}, Instant-NSR~\cite{zhao2022human} suffers from severe artifacts due to the heavy reliance on geometry. Volume rendering methods such as NeuS2~\cite{wang2023neus2} and HumanRF~\cite{isik2023humanrf} produce blurry results, over-smoothing on high-frequency details. Meanwhile, Dynamic 3D Gaussians~\cite{luiten2023dynamic} loses the advantages of 3DGS~\cite{kerbl20233d} due to fixed appearance attributes, failing to recover detailed appearance and view-dependency.
In contrast, HiFi4G surpasses these existing methods by merging 3D Gaussian representation with keyframe-update-based non-rigid tracking, providing detailed and high-quality human performance rendering.
For quantitative comparison, we evaluate each method on three 200-frame sequences from our dataset. We use various metrics, including PSNR, SSIM, LPIPS, the temporal metric VMAF~\cite{2016Toward}, and per-frame storage.
As seen in Tab.~\ref{table:Qualitativecomparison}, HiFi4G surpasses other methods in both quality and storage.
Note that our compression strategy significantly reduces per-frame storage requirements without compromising quality.
Remarkably, even on the VMAF metric~\cite{2016Toward}, which evaluates the perceptual quality and temporal consistency, our explicit method outperforms HumanRF which benefits from the inherent smoothness of the MLP.

\begin{table}[t]
	\begin{center}
		\centering
		\vspace{-10pt}
        \caption{
        Quantitative comparison of rendering results. Green and yellow cell colors indicate the best and the second-best results. %
        }
		\resizebox{0.47\textwidth}{!}{
			\begin{tabular}{l|ccccc}
				\hline
				Method   &  PSNR $\uparrow$ & SSIM $\uparrow$ & LPIPS $\downarrow$ & VMAF$\uparrow$ & Per-frame Storage(MB) $\downarrow$ \\
				\hline
                    Instant-NSR~\cite{zhao2022human} & 29.385 & 0.958 & 0.0370 & 68.309 & 11.63 \\
				NeuS2~\cite{wang2023neus2}\qquad\qquad & 32.952 & 0.961 & 0.0682 & 79.102 & 24.16 \\
			    HumanRF~\cite{isik2023humanrf}     & 31.174 & 0.977 & 0.0298 & 80.942 & 11.38 \\
				Dynamic 3D Gaussians~\cite{luiten2023dynamic}  & 30.244 & 0.965 & 0.0847 & 52.224 & \secondBestCellColor{4.523} \\
    		\hline

                Ours(Before Compression)  & \bestCellColor{\textbf{36.205}} & \bestCellColor{\textbf{0.989}} & \bestCellColor{\textbf{0.0184}} & \bestCellColor{\textbf{85.127}} & \textbf{43.42}\\
                Ours(After Compression) & \secondBestCellColor{\textbf{35.788}} & \secondBestCellColor{\textbf{0.986}} & \secondBestCellColor{\textbf{0.0214}} & \secondBestCellColor{\textbf{84.312}} & \bestCellColor{\textbf{1.648}}\\
				\hline
			\end{tabular}
		}
		\label{table:Qualitativecomparison}
    
    		\vspace{-25pt}
	\end{center}
\end{table}
\subsection{Ablation Study} \label{sec:abla} 

\begin{figure}[t] 
	\begin{center} 
		\includegraphics[width=1\linewidth]{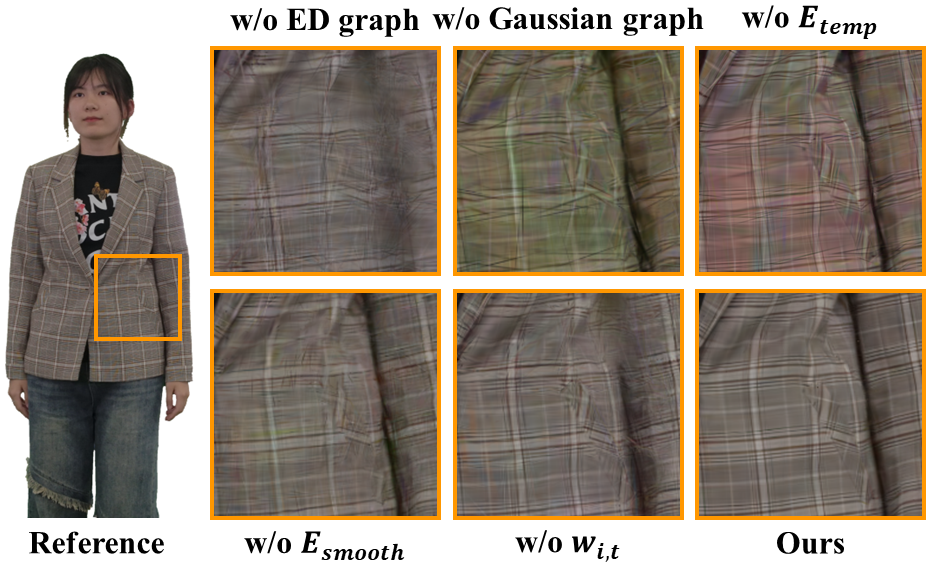}
	\end{center} 
	\vspace{-10pt}
	\caption{Qualitative evaluation of compact 4D Gaussians.} 
	\label{fig:term}
	\vspace{-10pt}
\end{figure}

\myparagraph{Compact 4D Gaussians.} We conduct a qualitative ablation on the dual-graph and the regularization term to assess their impact on post-compression rendering results.
As shown in Fig.~\ref{fig:term}, the removal of the coarse ED-graph prior typically causes severe artifacts. Excluding the Gaussian graph often results in significant precision loss and unnatural rendering.
Regarding regularizers, the omission of $E_{\mathrm{temp}}$ usually triggers unrealistic artifacts post-compression. Meanwhile, the absence of $E_{\mathrm{smooth}}$ produces blurry results, with both leading to flickering in the video. Additionally, to evaluate the impact of the adaptive weight $w_{i,t}$, we replace it with a fixed weight of 0.1. This adjustment generally leads to noticeable blurriness, especially in areas with significant movement.
In contrast, our full pipeline generates spatially and temporally compact 4D Gaussians, maintaining high-fidelity rendering even after compression. 
The quantitative results are as demonstrated in Tab.~\ref{table:term}, in which our full approach achieves the highest accuracy.

\begin{figure}[t] 
	\begin{center} 
		\includegraphics[width=1\linewidth]{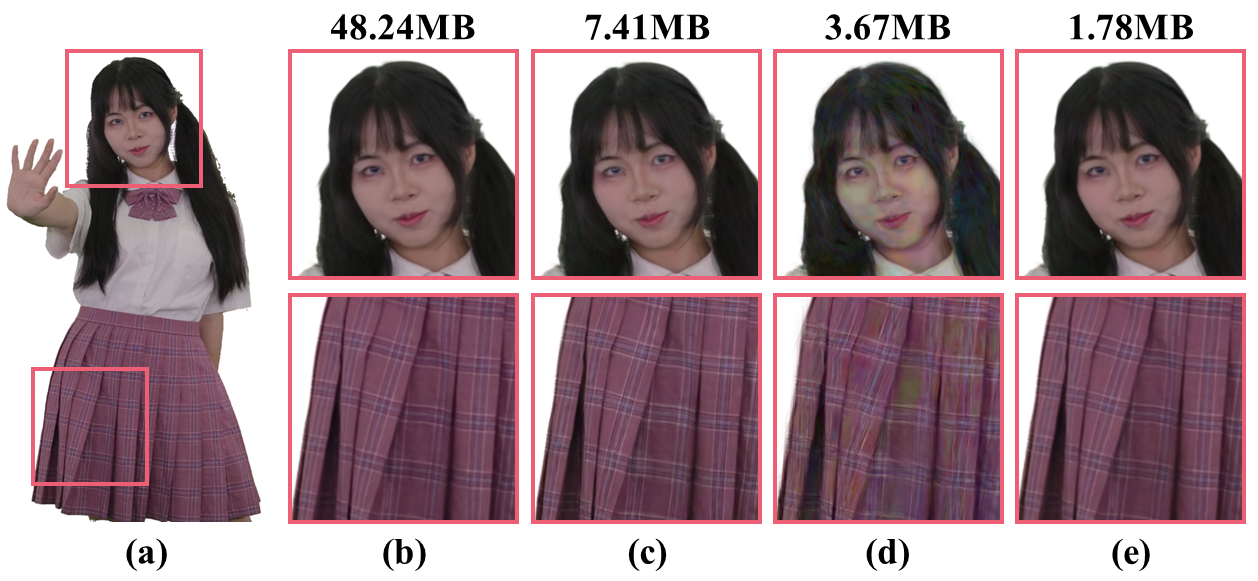}
	\end{center} 
	\vspace{-14pt}
	\caption{Qualitative evaluation of our residual strategy. (a) Reference image; (b) 4D Gaussians results before compression; (c) Per-frame encoding using high-bit quantization without residual; (d) Per-frame encoding using low-bit quantization without residual; (e) Ours results using low-bit quantization with residual.} 
	\label{fig:compress_eval}
	\vspace{-10pt}
\end{figure}

\myparagraph{Residual Compensation.}
As illustrated in Fig.~\ref{fig:compress_eval} (b), we allocate 48.24MB of storage for the 4D Gaussians of each frame before compression. Applying high-bit quantization (0-bit for motion and 9-bit for appearance) without residual compensation results in a storage requirement of 7.41MB, as shown in Fig.~\ref{fig:compress_eval} (c). Using low-bit quantization (11-bit for motion and 7-bit for appearance), again without residual compensation, reduces storage to 3.67MB but compromises rendering quality, as illustrated in Fig.~\ref{fig:compress_eval} (d).
In contrast, applying the same low-bit quantization but with residual compensation significantly reduces storage needs to under 2MB per frame while maintaining the same level of rendering quality, as shown in Fig.~\ref{fig:compress_eval} (e).

\begin{figure}[t] 
	\begin{center} 
		\includegraphics[width=1\linewidth]{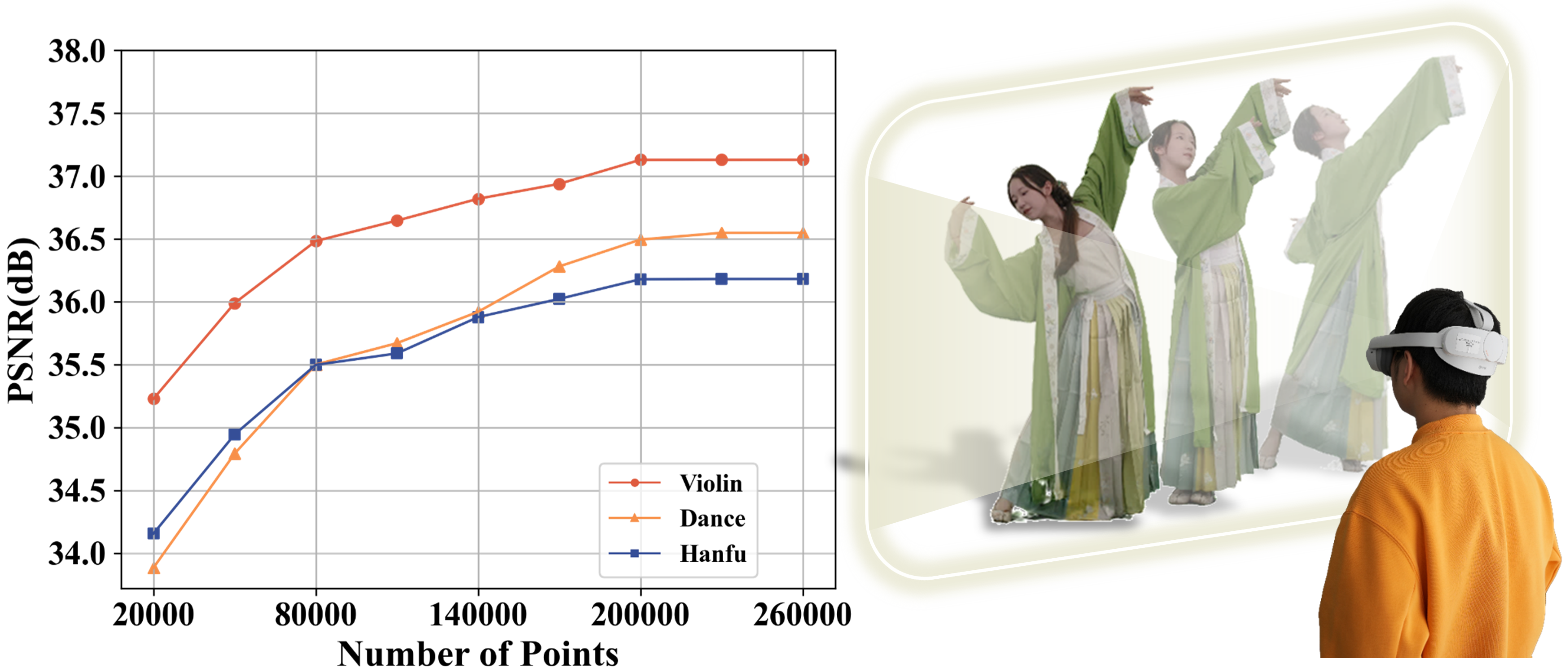} 
	\end{center} 
	\vspace{-10pt}
	\caption{Evaluation of 4D Gaussians number. With $\sim$200,000 4D Gaussians, HiFi4G achieves high-fidelity human performance rendering, suitable for integration in VR applications.} 
	\label{fig:point_number}
	\vspace{-10pt}
\end{figure}

\myparagraph{The Number of 4D Gaussians.}
We assess the impact of changing the number of 4D Gaussians on the quality of results across three sequences. As depicted in Fig.~\ref{fig:point_number}, using 200,000 4D Gaussians is adequate for generating high-quality results. This amount enables effective compression to less than 2MB, supporting immersive viewing on diverse platforms, including VR and AR.

\myparagraph{Run-time Evaluation of Each Step.}
As shown in Tab.~\ref{table:running-time}, we also provide the runtime for each step on a PC with an Nvidia GeForce RTX3090 GPU, which includes both the preprocessing and training stages. Our method can generate 4D assets efficiently, taking less than 7 minutes per frame.
\begin{table}[t]
    \Huge
	\begin{center}
		\centering
		\caption{Quantitative evaluation of compact 4D Gaussians.}
		\vspace{-10pt}
		\label{table:term}
		\resizebox{0.45\textwidth}{!}{
			\begin{tabular}{l|cccc}
				\hline
				       & PSNR $\uparrow $
				& SSIM  $\uparrow $ & LPIPS $\downarrow$ & VMAF $\uparrow$   \\
				\hline
				w/o ED graph       & 29.142            & 0.9534 & 0.0662 &  69.724     \\
				w/o Gaussian graph      & 31.185     & 0.9541 & 0.0586  &76.873
    \\
				\hline
				\hline
                 w/o $E_{\text{temp}}$        & 33.555   & 0.9661 & 0.0496 &  76.308 \\
				w/o $E_{\text{smooth}}$      & 33.889            & 0.9657   &0.0518 & 81.944        \\
				
    		w/o $w_{i,t}$      & 33.577     &   0.9678 &  0.0425  &  81.236  \\

				\hline
                    Ours &35.085 & 0.9828 &0.0219& 83.133\\
                    \hline
			\end{tabular}
		}
		\vspace{-20pt}
	\end{center}
\end{table}
\begin{table}[t]\tiny
	\begin{center}
		\centering
		\caption{Run-time evaluation of each step.}
		\vspace{-8pt}
		\label{table:running-time}
		\resizebox{0.4\textwidth}{!}{
			\begin{tabular}{l|cccc}
				\hline
				     Procedure  &  Time\\
				\hline
			    	 Background matting   \quad\quad\quad \quad\quad\quad\quad    &\quad\quad\quad $\sim$ 1 min   \quad\quad\quad\quad\\
					 Meshing                 & $\sim$ 1 min     \\
                \hline
                    non-rigid tracking            &  $\sim$ 100 ms \\

			       4D Gaussians Optimization   & $\sim$ 4 mins\\
					 4D Gaussians Compression     & $\sim$ 100 ms \\
				\hline
			\end{tabular}
		}
		\vspace{-20pt}
	\end{center}
\end{table}
\subsection{Limitation}
Although HiFi4G achieves high-fidelity 4D human performance rendering via compact Gaussian Splatting, it still has some limitations. First, HiFi4G heavily relies on segmentation, poor segmentation can lead to significant artifacts, especially in scenes with human-object interactions. Moreover, our method necessitates per-frame reconstruction and mesh tracking, which presents an interesting direction in exploring a more synergistic relationship between tracking and rendering. Even though HiFi4G is efficient in generating 4D assets, the Gaussian optimization process still requires several minutes, forming a major bottleneck. Accelerating this training process is vital for future research.
Additionally, the current dependence of 4D Gaussian on fast GPU sorting limits the deployment of HiFi4G on web viewers and mobile devices.

\section {Conclusion} 
We have presented an explicit and compact Gaussian-based approach for 4D human performance rendering from RGB inputs. By bridging 3D Gaussian Splatting with non-rigid tracking, our approach achieves high-fidelity rendering results, outperforming previous methods in terms of quality, efficiency, and storage.  Our dual-graph mechanism provides sufficient non-rigid motion priors in a keyframe-based manner, while our Gaussian optimization scheme with novel regularization designs effectively ensures spatial-temporal consistency of the 4D Gaussian Splatting. We also demonstrate the compactness of our representation with a companion compression scheme which substantially reduces storage requirements. Our experimental results further demonstrate the effectiveness of our approach for delivering lifelike human performances. 
With its explicit and compact characteristics, we believe our approach makes a solid step forward to faithfully recording and providing immersive experiences of human performances on various platforms like VR headsets.

{\small
\bibliographystyle{ieeenat_fullname}
\bibliography{reference}

\begin{thebibliography}{80}
\providecommand{\natexlab}[1]{#1}
\providecommand{\url}[1]{\texttt{#1}}
\expandafter\ifx\csname urlstyle\endcsname\relax
  \providecommand{\doi}[1]{doi: #1}\else
  \providecommand{\doi}{doi: \begingroup \urlstyle{rm}\Url}\fi

\bibitem[Bozic et~al.(2021)Bozic, Palafox, Zollhofer, Thies, Dai, and
  Nie{\ss}ner]{bozic2021neural}
Aljaz Bozic, Pablo Palafox, Michael Zollhofer, Justus Thies, Angela Dai, and
  Matthias Nie{\ss}ner.
\newblock Neural deformation graphs for globally-consistent non-rigid
  reconstruction.
\newblock In \emph{Proceedings of the IEEE/CVF Conference on Computer Vision
  and Pattern Recognition}, pages 1450--1459, 2021.

\bibitem[Burov et~al.(2021)Burov, Nie{\ss}ner, and Thies]{burov2021dsfn}
Andrei Burov, Matthias Nie{\ss}ner, and Justus Thies.
\newblock Dynamic surface function networks for clothed human bodies, 2021.

\bibitem[Cai et~al.(2022)Cai, Feng, Feng, Wang, and Zhang]{Cai2022NDR}
Hongrui Cai, Wanquan Feng, Xuetao Feng, Yan Wang, and Juyong Zhang.
\newblock Neural surface reconstruction of dynamic scenes with monocular rgb-d
  camera.
\newblock In \emph{Thirty-sixth Conference on Neural Information Processing
  Systems (NeurIPS)}, 2022.

\bibitem[Cao and Johnson(2023)]{cao2023hexplane}
Ang Cao and Justin Johnson.
\newblock Hexplane: A fast representation for dynamic scenes.
\newblock In \emph{Proceedings of the IEEE/CVF Conference on Computer Vision
  and Pattern Recognition}, pages 130--141, 2023.

\bibitem[Cao et~al.(2017)Cao, Simon, Wei, and Sheikh]{OpenPose}
Zhe Cao, Tomas Simon, Shih-En Wei, and Yaser Sheikh.
\newblock Realtime multi-person 2d pose estimation using part affinity fields.
\newblock In \emph{Computer Vision and Pattern Recognition (CVPR)}, 2017.

\bibitem[Carandini et~al.(2005)Carandini, Demb, Mante, Tolhurst, Dan,
  Olshausen, Gallant, and Rust]{carandini2005we}
Matteo Carandini, Jonathan~B Demb, Valerio Mante, David~J Tolhurst, Yang Dan,
  Bruno~A Olshausen, Jack~L Gallant, and Nicole~C Rust.
\newblock Do we know what the early visual system does?
\newblock \emph{Journal of Neuroscience}, 25\penalty0 (46):\penalty0
  10577--10597, 2005.

\bibitem[Chen et~al.(2022)Chen, Xu, Geiger, Yu, and Su]{chen2022tensorf}
Anpei Chen, Zexiang Xu, Andreas Geiger, Jingyi Yu, and Hao Su.
\newblock Tensorf: Tensorial radiance fields.
\newblock In \emph{European Conference on Computer Vision}, pages 333--350.
  Springer, 2022.

\bibitem[Chen et~al.(2021)Chen, Zheng, Black, Hilliges, and
  Geiger]{chen2021snarf}
Xu Chen, Yufeng Zheng, Michael~J Black, Otmar Hilliges, and Andreas Geiger.
\newblock Snarf: Differentiable forward skinning for animating non-rigid neural
  implicit shapes.
\newblock In \emph{Proceedings of the IEEE/CVF International Conference on
  Computer Vision}, pages 11594--11604, 2021.

\bibitem[Collet et~al.(2015)Collet, Chuang, Sweeney, Gillett, Evseev,
  Calabrese, Hoppe, Kirk, and Sullivan]{collet2015high}
Alvaro Collet, Ming Chuang, Pat Sweeney, Don Gillett, Dennis Evseev, David
  Calabrese, Hugues Hoppe, Adam Kirk, and Steve Sullivan.
\newblock High-quality streamable free-viewpoint video.
\newblock \emph{ACM Transactions on Graphics (TOG)}, 34\penalty0 (4):\penalty0
  69, 2015.

\bibitem[Dou et~al.(2016)Dou, Khamis, Degtyarev, Davidson, Fanello, Kowdle,
  Escolano, Rhemann, Kim, Taylor, et~al.]{dou2016fusion4d}
Mingsong Dou, Sameh Khamis, Yury Degtyarev, Philip Davidson, Sean~Ryan Fanello,
  Adarsh Kowdle, Sergio~Orts Escolano, Christoph Rhemann, David Kim, Jonathan
  Taylor, et~al.
\newblock Fusion4d: Real-time performance capture of challenging scenes.
\newblock \emph{ACM Transactions on Graphics (ToG)}, 35\penalty0 (4):\penalty0
  1--13, 2016.

\bibitem[Dou et~al.(2017)Dou, Davidson, Fanello, Khamis, Kowdle, Rhemann,
  Tankovich, and Izadi]{motion2fusion}
Mingsong Dou, Philip Davidson, Sean~Ryan Fanello, Sameh Khamis, Adarsh Kowdle,
  Christoph Rhemann, Vladimir Tankovich, and Shahram Izadi.
\newblock Motion2fusion: Real-time volumetric performance capture.
\newblock \emph{ACM Trans. Graph.}, 36\penalty0 (6):\penalty0 246:1--246:16,
  2017.

\bibitem[Duda(2013)]{duda2013asymmetric}
Jarek Duda.
\newblock Asymmetric numeral systems: entropy coding combining speed of huffman
  coding with compression rate of arithmetic coding.
\newblock \emph{arXiv preprint arXiv:1311.2540}, 2013.

\bibitem[Fridovich-Keil et~al.(2023)Fridovich-Keil, Meanti, Warburg, Recht, and
  Kanazawa]{fridovich2023k}
Sara Fridovich-Keil, Giacomo Meanti, Frederik~Rahb{\ae}k Warburg, Benjamin
  Recht, and Angjoo Kanazawa.
\newblock K-planes: Explicit radiance fields in space, time, and appearance.
\newblock In \emph{Proceedings of the IEEE/CVF Conference on Computer Vision
  and Pattern Recognition}, pages 12479--12488, 2023.

\bibitem[Gafni et~al.(2021)Gafni, Thies, Zollh{\"o}fer, and
  Nie{\ss}ner]{Gafni_2021_CVPR}
Guy Gafni, Justus Thies, Michael Zollh{\"o}fer, and Matthias Nie{\ss}ner.
\newblock Dynamic neural radiance fields for monocular 4d facial avatar
  reconstruction.
\newblock In \emph{Proceedings of the IEEE/CVF Conference on Computer Vision
  and Pattern Recognition (CVPR)}, pages 8649--8658, 2021.

\bibitem[Guo et~al.(2015)Guo, Xu, Wang, Liu, and Dai]{guo2015robust}
Kaiwen Guo, Feng Xu, Yangang Wang, Yebin Liu, and Qionghai Dai.
\newblock {Robust Non-Rigid Motion Tracking and Surface Reconstruction Using L0
  Regularization}.
\newblock In \emph{Proceedings of the IEEE International Conference on Computer
  Vision}, pages 3083--3091, 2015.

\bibitem[Guo et~al.(2017)Guo, Xu, Yu, Liu, Dai, and Liu]{guo2017real}
Kaiwen Guo, Feng Xu, Tao Yu, Xiaoyang Liu, Qionghai Dai, and Yebin Liu.
\newblock Real-time geometry, albedo and motion reconstruction using a single
  rgbd camera.
\newblock \emph{ACM Transactions on Graphics (TOG)}, 2017.

\bibitem[He et~al.(2021)He, Pang, Chen, Liang, Wu, Ma, and Xu]{Challencap2021}
Yannan He, Anqi Pang, Xin Chen, Han Liang, Minye Wu, Yuexin Ma, and Lan Xu.
\newblock Challencap: Monocular 3d capture of challenging human performances
  using multi-modal references.
\newblock In \emph{Proceedings of the IEEE/CVF Conference on Computer Vision
  and Pattern Recognition (CVPR)}, 2021.

\bibitem[Hu et~al.(2023)Hu, Wang, Ma, Yang, Gao, Liu, and Ma]{hu2023tri}
Wenbo Hu, Yuling Wang, Lin Ma, Bangbang Yang, Lin Gao, Xiao Liu, and Yuewen Ma.
\newblock Tri-miprf: Tri-mip representation for efficient anti-aliasing neural
  radiance fields.
\newblock In \emph{Proceedings of the IEEE/CVF International Conference on
  Computer Vision}, pages 19774--19783, 2023.

\bibitem[I\c{s}{\i}k et~al.(2023)I\c{s}{\i}k, Rünz, Georgopoulos, Khakhulin,
  Starck, Agapito, and Nießner]{isik2023humanrf}
Mustafa I\c{s}{\i}k, Martin Rünz, Markos Georgopoulos, Taras Khakhulin,
  Jonathan Starck, Lourdes Agapito, and Matthias Nießner.
\newblock Humanrf: High-fidelity neural radiance fields for humans in motion.
\newblock \emph{ACM Transactions on Graphics (TOG)}, 42\penalty0 (4):\penalty0
  1--12, 2023.

\bibitem[Jiang et~al.(2022{\natexlab{a}})Jiang, Habermann, Golyanik, and
  Theobalt]{jiang2022hifecap}
Yue Jiang, Marc Habermann, Vladislav Golyanik, and Christian Theobalt.
\newblock Hifecap: Monocular high-fidelity and expressive capture of human
  performances.
\newblock In \emph{BMVC}, 2022{\natexlab{a}}.

\bibitem[Jiang et~al.(2022{\natexlab{b}})Jiang, Jiang, Sun, Su, Guo, Wu, Yu,
  and Xu]{jiang2022neuralhofusion}
Yuheng Jiang, Suyi Jiang, Guoxing Sun, Zhuo Su, Kaiwen Guo, Minye Wu, Jingyi
  Yu, and Lan Xu.
\newblock Neuralhofusion: Neural volumetric rendering under human-object
  interactions.
\newblock In \emph{Proceedings of the IEEE/CVF Conference on Computer Vision
  and Pattern Recognition}, pages 6155--6165, 2022{\natexlab{b}}.

\bibitem[Jiang et~al.(2023)Jiang, Yao, Su, Shen, Luo, and Xu]{jiang2023instant}
Yuheng Jiang, Kaixin Yao, Zhuo Su, Zhehao Shen, Haimin Luo, and Lan Xu.
\newblock Instant-nvr: Instant neural volumetric rendering for human-object
  interactions from monocular rgbd stream.
\newblock In \emph{Proceedings of the IEEE/CVF Conference on Computer Vision
  and Pattern Recognition}, pages 595--605, 2023.

\bibitem[Joo et~al.(2018)Joo, Simon, and Sheikh]{TotalCapture}
Hanbyul Joo, Tomas Simon, and Yaser Sheikh.
\newblock Total capture: A 3d deformation model for tracking faces, hands, and
  bodies.
\newblock In \emph{The IEEE Conference on Computer Vision and Pattern
  Recognition (CVPR)}, 2018.

\bibitem[Kerbl et~al.(2023)Kerbl, Kopanas, Leimk{\"u}hler, and
  Drettakis]{kerbl20233d}
Bernhard Kerbl, Georgios Kopanas, Thomas Leimk{\"u}hler, and George Drettakis.
\newblock 3d gaussian splatting for real-time radiance field rendering.
\newblock \emph{ACM Transactions on Graphics (ToG)}, 42\penalty0 (4):\penalty0
  1--14, 2023.

\bibitem[Li et~al.(2022{\natexlab{a}})Li, Tanke, Vo, Zollh{\"o}fer, Gall,
  Kanazawa, and Lassner]{li2022tava}
Ruilong Li, Julian Tanke, Minh Vo, Michael Zollh{\"o}fer, J{\"u}rgen Gall,
  Angjoo Kanazawa, and Christoph Lassner.
\newblock Tava: Template-free animatable volumetric actors.
\newblock In \emph{European Conference on Computer Vision}, pages 419--436.
  Springer, 2022{\natexlab{a}}.

\bibitem[Li et~al.(2021{\natexlab{a}})Li, Takehara, Taketomi, Zheng, and
  Nie{\ss}ner]{li20214dcomplete}
Yang Li, Hikari Takehara, Takafumi Taketomi, Bo Zheng, and Matthias
  Nie{\ss}ner.
\newblock 4dcomplete: Non-rigid motion estimation beyond the observable
  surface.
\newblock In \emph{Proceedings of the IEEE/CVF International Conference on
  Computer Vision}, 2021{\natexlab{a}}.

\bibitem[Li et~al.(2016)Li, Aaron, Katsavounidis, Moorthy, Manohara,
  et~al.]{2016Toward}
Zhi Li, Anne Aaron, Ioannis Katsavounidis, Anush Moorthy, Megha Manohara,
  et~al.
\newblock Toward a practical perceptual video quality metric.
\newblock \emph{The Netflix Tech Blog}, 6\penalty0 (2):\penalty0 2, 2016.

\bibitem[Li et~al.(2021{\natexlab{b}})Li, Yu, Zheng, Guo, and
  Liu]{li2021posefusion}
Zhe Li, Tao Yu, Zerong Zheng, Kaiwen Guo, and Yebin Liu.
\newblock Posefusion: Pose-guided selective fusion for single-view human
  volumetric capture.
\newblock In \emph{IEEE Conference on Computer Vision and Pattern Recognition},
  2021{\natexlab{b}}.

\bibitem[Li et~al.(2022{\natexlab{b}})Li, Zheng, Zhang, Ji, and
  Liu]{li2022avatarcap}
Zhe Li, Zerong Zheng, Hongwen Zhang, Chaonan Ji, and Yebin Liu.
\newblock Avatarcap: Animatable avatar conditioned monocular human volumetric
  capture.
\newblock In \emph{ECCV}, 2022{\natexlab{b}}.

\bibitem[Liang and Liang(2022)]{liang2022transpcc}
Zujie Liang and Fan Liang.
\newblock Transpcc: Towards deep point cloud compression via transformers.
\newblock In \emph{Proceedings of the 2022 International Conference on
  Multimedia Retrieval}, pages 1--5, 2022.

\bibitem[Lin et~al.(2022)Lin, Peng, Xu, Yan, Shuai, Bao, and
  Zhou]{lin2022efficient}
Haotong Lin, Sida Peng, Zhen Xu, Yunzhi Yan, Qing Shuai, Hujun Bao, and Xiaowei
  Zhou.
\newblock Efficient neural radiance fields for interactive free-viewpoint
  video.
\newblock In \emph{SIGGRAPH Asia Conference Proceedings}, 2022.

\bibitem[Lin et~al.(2023)Lin, Peng, Xu, Xie, He, Bao, and Zhou]{lin2023im4d}
Haotong Lin, Sida Peng, Zhen Xu, Tao Xie, Xingyi He, Hujun Bao, and Xiaowei
  Zhou.
\newblock Im4d: High-fidelity and real-time novel view synthesis for dynamic
  scenes.
\newblock \emph{arXiv preprint arXiv:2310.08585}, 2023.

\bibitem[Lin et~al.(2021)Lin, Ryabtsev, Sengupta, Curless, Seitz, and
  Kemelmacher-Shlizerman]{lin2021real}
Shanchuan Lin, Andrey Ryabtsev, Soumyadip Sengupta, Brian~L Curless, Steven~M
  Seitz, and Ira Kemelmacher-Shlizerman.
\newblock Real-time high-resolution background matting.
\newblock In \emph{Proceedings of the IEEE/CVF Conference on Computer Vision
  and Pattern Recognition}, pages 8762--8771, 2021.

\bibitem[Liu et~al.(2020)Liu, Xu, Habermann, Zollhöfer, Bernard, Kim, Wang,
  and Theobalt]{liu2020NeuralHumanRendering}
Lingjie Liu, Weipeng Xu, Marc Habermann, Michael Zollhöfer, Florian Bernard,
  Hyeongwoo Kim, Wenping Wang, and Christian Theobalt.
\newblock Neural human video rendering by learning dynamic textures and
  rendering-to-video translation.
\newblock \emph{IEEE Transactions on Visualization and Computer Graphics},
  PP:\penalty0 1--1, 2020.

\bibitem[Loper et~al.(2015)Loper, Mahmood, Romero, Pons-Moll, and
  Black]{SMPL2015}
Matthew Loper, Naureen Mahmood, Javier Romero, Gerard Pons-Moll, and Michael~J.
  Black.
\newblock Smpl: A skinned multi-person linear model.
\newblock \emph{ACM Trans. Graph.}, 34\penalty0 (6):\penalty0 248:1--248:16,
  2015.

\bibitem[Luiten et~al.(2023)Luiten, Kopanas, Leibe, and
  Ramanan]{luiten2023dynamic}
Jonathon Luiten, Georgios Kopanas, Bastian Leibe, and Deva Ramanan.
\newblock Dynamic 3d gaussians: Tracking by persistent dynamic view synthesis.
\newblock \emph{arXiv preprint arXiv:2308.09713}, 2023.

\bibitem[Luo et~al.(2022)Luo, Xu, Jiang, Zhou, Qiu, Zhang, Yang, Xu, and
  Yu]{luo2022artemis}
Haimin Luo, Teng Xu, Yuheng Jiang, Chenglin Zhou, Qiwei Qiu, Yingliang Zhang,
  Wei Yang, Lan Xu, and Jingyi Yu.
\newblock Artemis: Articulated neural pets with appearance and motion
  synthesis.
\newblock \emph{ACM Trans. Graph.}, 41\penalty0 (4), 2022.

\bibitem[Ma et~al.(2022)Ma, Yang, Black, and Tang]{SkiRT:3DV:2022}
Qianli Ma, Jinlong Yang, Michael~J. Black, and Siyu Tang.
\newblock Neural point-based shape modeling of humans in challenging clothing.
\newblock In \emph{2022 International Conference on 3D Vision (3DV)}, 2022.

\bibitem[Mildenhall et~al.(2020)Mildenhall, Srinivasan, Tancik, Barron,
  Ramamoorthi, and Ng]{nerf}
Ben Mildenhall, Pratul~P. Srinivasan, Matthew Tancik, Jonathan~T. Barron, Ravi
  Ramamoorthi, and Ren Ng.
\newblock Nerf: Representing scenes as neural radiance fields for view
  synthesis.
\newblock In \emph{Computer Vision -- ECCV 2020}, pages 405--421, Cham, 2020.
  Springer International Publishing.

\bibitem[M\"uller et~al.(2022)M\"uller, Evans, Schied, and
  Keller]{muller2022instant}
Thomas M\"uller, Alex Evans, Christoph Schied, and Alexander Keller.
\newblock Instant neural graphics primitives with a multiresolution hash
  encoding.
\newblock \emph{ACM Trans. Graph.}, 41\penalty0 (4):\penalty0 102:1--102:15,
  2022.

\bibitem[Nadenau et~al.(2003)Nadenau, Reichel, and Kunt]{nadenau2003wavelet}
Marcus~J Nadenau, Julien Reichel, and Murat Kunt.
\newblock Wavelet-based color image compression: exploiting the contrast
  sensitivity function.
\newblock \emph{IEEE Transactions on image processing}, 12\penalty0
  (1):\penalty0 58--70, 2003.

\bibitem[Newcombe et~al.(2015)Newcombe, Fox, and
  Seitz]{newcombe2015dynamicfusion}
Richard~A Newcombe, Dieter Fox, and Steven~M Seitz.
\newblock Dynamicfusion: Reconstruction and tracking of non-rigid scenes in
  real-time.
\newblock In \emph{Proceedings of the IEEE conference on computer vision and
  pattern recognition}, pages 343--352, 2015.

\bibitem[Palafox et~al.(2021)Palafox, Bo{\v{z}}i{\v{c}}, Thies, Nie{\ss}ner,
  and Dai]{palafox2021npms}
Pablo Palafox, Alja{\v{z}} Bo{\v{z}}i{\v{c}}, Justus Thies, Matthias
  Nie{\ss}ner, and Angela Dai.
\newblock Npms: Neural parametric models for 3d deformable shapes.
\newblock In \emph{Proceedings of the IEEE/CVF International Conference on
  Computer Vision}, pages 12695--12705, 2021.

\bibitem[Park et~al.(2021{\natexlab{a}})Park, Sinha, Barron, Bouaziz, Goldman,
  Seitz, and Martin-Brualla]{park2021nerfies}
Keunhong Park, Utkarsh Sinha, Jonathan~T Barron, Sofien Bouaziz, Dan~B Goldman,
  Steven~M Seitz, and Ricardo Martin-Brualla.
\newblock Nerfies: Deformable neural radiance fields.
\newblock In \emph{Proceedings of the IEEE/CVF International Conference on
  Computer Vision}, pages 5865--5874, 2021{\natexlab{a}}.

\bibitem[Park et~al.(2021{\natexlab{b}})Park, Sinha, Hedman, Barron, Bouaziz,
  Goldman, Martin-Brualla, and Seitz]{park2021hypernerf}
Keunhong Park, Utkarsh Sinha, Peter Hedman, Jonathan~T. Barron, Sofien Bouaziz,
  Dan~B Goldman, Ricardo Martin-Brualla, and Steven~M. Seitz.
\newblock Hypernerf: A higher-dimensional representation for topologically
  varying neural radiance fields.
\newblock \emph{ACM Trans. Graph.}, 40\penalty0 (6), 2021{\natexlab{b}}.

\bibitem[Peng et~al.(2021{\natexlab{a}})Peng, Dong, Wang, Zhang, Shuai, Zhou,
  and Bao]{peng2021animatable}
Sida Peng, Junting Dong, Qianqian Wang, Shangzhan Zhang, Qing Shuai, Xiaowei
  Zhou, and Hujun Bao.
\newblock Animatable neural radiance fields for modeling dynamic human bodies.
\newblock In \emph{Proceedings of the IEEE/CVF International Conference on
  Computer Vision}, pages 14314--14323, 2021{\natexlab{a}}.

\bibitem[Peng et~al.(2021{\natexlab{b}})Peng, Zhang, Xu, Wang, Shuai, Bao, and
  Zhou]{peng2021neural}
Sida Peng, Yuanqing Zhang, Yinghao Xu, Qianqian Wang, Qing Shuai, Hujun Bao,
  and Xiaowei Zhou.
\newblock Neural body: Implicit neural representations with structured latent
  codes for novel view synthesis of dynamic humans.
\newblock In \emph{Proceedings of the IEEE/CVF Conference on Computer Vision
  and Pattern Recognition}, pages 9054--9063, 2021{\natexlab{b}}.

\bibitem[Prokudin et~al.(2023)Prokudin, Ma, Raafat, Valentin, and
  Tang]{Prokudin_2023_ICCV}
Sergey Prokudin, Qianli Ma, Maxime Raafat, Julien Valentin, and Siyu Tang.
\newblock Dynamic point fields.
\newblock In \emph{Proceedings of the IEEE/CVF International Conference on
  Computer Vision (ICCV)}, pages 7964--7976, 2023.

\bibitem[Pumarola et~al.(2021)Pumarola, Corona, Pons-Moll, and
  Moreno-Noguer]{pumarola2020d}
Albert Pumarola, Enric Corona, Gerard Pons-Moll, and Francesc Moreno-Noguer.
\newblock D-nerf: Neural radiance fields for dynamic scenes.
\newblock In \emph{Proceedings of the IEEE/CVF Conference on Computer Vision
  and Pattern Recognition}, pages 10318--10327, 2021.

\bibitem[Quach et~al.(2019)Quach, Valenzise, and Dufaux]{quach2019learning}
Maurice Quach, Giuseppe Valenzise, and Frederic Dufaux.
\newblock Learning convolutional transforms for lossy point cloud geometry
  compression.
\newblock In \emph{2019 IEEE international conference on image processing
  (ICIP)}, pages 4320--4324. IEEE, 2019.

\bibitem[Quach et~al.(2020)Quach, Valenzise, and Dufaux]{quach2020improved}
Maurice Quach, Giuseppe Valenzise, and Frederic Dufaux.
\newblock Improved deep point cloud geometry compression.
\newblock In \emph{2020 IEEE 22nd International Workshop on Multimedia Signal
  Processing (MMSP)}, pages 1--6. IEEE, 2020.

\bibitem[Reiser et~al.(2023)Reiser, Szeliski, Verbin, Srinivasan, Mildenhall,
  Geiger, Barron, and Hedman]{reiser2023merf}
Christian Reiser, Rick Szeliski, Dor Verbin, Pratul Srinivasan, Ben Mildenhall,
  Andreas Geiger, Jon Barron, and Peter Hedman.
\newblock Merf: Memory-efficient radiance fields for real-time view synthesis
  in unbounded scenes.
\newblock \emph{ACM Transactions on Graphics (TOG)}, 42\penalty0 (4):\penalty0
  1--12, 2023.

\bibitem[Schnabel and Klein(2006)]{schnabel2006octree}
Ruwen Schnabel and Reinhard Klein.
\newblock Octree-based point-cloud compression.
\newblock \emph{PBG@ SIGGRAPH}, 3, 2006.

\bibitem[Schwarz et~al.(2018)Schwarz, Preda, Baroncini, Budagavi, Cesar, Chou,
  Cohen, Krivoku{\'c}a, Lasserre, Li, et~al.]{schwarz2018emerging}
Sebastian Schwarz, Marius Preda, Vittorio Baroncini, Madhukar Budagavi, Pablo
  Cesar, Philip~A Chou, Robert~A Cohen, Maja Krivoku{\'c}a, S{\'e}bastien
  Lasserre, Zhu Li, et~al.
\newblock Emerging mpeg standards for point cloud compression.
\newblock \emph{IEEE Journal on Emerging and Selected Topics in Circuits and
  Systems}, 9\penalty0 (1):\penalty0 133--148, 2018.

\bibitem[Shao et~al.(2023)Shao, Zheng, Tu, Liu, Zhang, and
  Liu]{shao2023tensor4d}
Ruizhi Shao, Zerong Zheng, Hanzhang Tu, Boning Liu, Hongwen Zhang, and Yebin
  Liu.
\newblock Tensor4d: Efficient neural 4d decomposition for high-fidelity dynamic
  reconstruction and rendering.
\newblock In \emph{Proceedings of the IEEE/CVF Conference on Computer Vision
  and Pattern Recognition}, pages 16632--16642, 2023.

\bibitem[Shen et~al.(2023)Shen, Guo, Kaufmann, Zarate, Valentin, Song, and
  Hilliges]{shen2023x}
Kaiyue Shen, Chen Guo, Manuel Kaufmann, Juan~Jose Zarate, Julien Valentin, Jie
  Song, and Otmar Hilliges.
\newblock X-avatar: Expressive human avatars.
\newblock In \emph{Proceedings of the IEEE/CVF Conference on Computer Vision
  and Pattern Recognition}, pages 16911--16921, 2023.

\bibitem[Song et~al.(2023)Song, Chen, Li, Chen, Chen, Yuan, Xu, and
  Geiger]{song2023nerfplayer}
Liangchen Song, Anpei Chen, Zhong Li, Zhang Chen, Lele Chen, Junsong Yuan, Yi
  Xu, and Andreas Geiger.
\newblock Nerfplayer: A streamable dynamic scene representation with decomposed
  neural radiance fields.
\newblock \emph{IEEE Transactions on Visualization and Computer Graphics},
  29\penalty0 (5):\penalty0 2732--2742, 2023.

\bibitem[Su et~al.(2020)Su, Xu, Zheng, Yu, Liu, and Fang]{robustfusion}
Zhuo Su, Lan Xu, Zerong Zheng, Tao Yu, Yebin Liu, and Lu Fang.
\newblock Robustfusion: Human volumetric capture with data-driven visual cues
  using a rgbd camera.
\newblock In \emph{Computer Vision -- ECCV 2020}, pages 246--264, Cham, 2020.
  Springer International Publishing.

\bibitem[Su et~al.(2022)Su, Xu, Zhong, Li, Deng, Quan, and
  Fang]{su2022robustfusionPlus}
Zhuo Su, Lan Xu, Dawei Zhong, Zhong Li, Fan Deng, Shuxue Quan, and Lu Fang.
\newblock Robustfusion: Robust volumetric performance reconstruction under
  human-object interactions from monocular rgbd stream.
\newblock \emph{IEEE Transactions on Pattern Analysis and Machine
  Intelligence}, 2022.

\bibitem[Sumner et~al.(2007)Sumner, Schmid, and Pauly]{sumner2007embedded}
Robert~W Sumner, Johannes Schmid, and Mark Pauly.
\newblock Embedded deformation for shape manipulation.
\newblock \emph{ACM Transactions on Graphics (TOG)}, 26\penalty0 (3):\penalty0
  80, 2007.

\bibitem[Sun et~al.(2021)Sun, Chen, Chen, Pang, Lin, Jiang, Xu, Wang, and
  Yu]{sun2021HOI-FVV}
Guoxing Sun, Xin Chen, Yizhang Chen, Anqi Pang, Pei Lin, Yuheng Jiang, Lan Xu,
  Jingya Wang, and Jingyi Yu.
\newblock Neural free-viewpoint performance rendering under complex
  human-object interactions.
\newblock In \emph{Proceedings of the 29th ACM International Conference on
  Multimedia}, 2021.

\bibitem[Suo et~al.(2021)Suo, Jiang, Lin, Zhang, Wu, Guo, and
  Xu]{suo2021neuralhumanfvv}
Xin Suo, Yuheng Jiang, Pei Lin, Yingliang Zhang, Minye Wu, Kaiwen Guo, and Lan
  Xu.
\newblock Neuralhumanfvv: Real-time neural volumetric human performance
  rendering using rgb cameras.
\newblock In \emph{Proceedings of the IEEE/CVF Conference on Computer Vision
  and Pattern Recognition}, pages 6226--6237, 2021.

\bibitem[Thanou et~al.(2016)Thanou, Chou, and Frossard]{thanou2016graph}
Dorina Thanou, Philip~A Chou, and Pascal Frossard.
\newblock Graph-based compression of dynamic 3d point cloud sequences.
\newblock \emph{IEEE Transactions on Image Processing}, 25\penalty0
  (4):\penalty0 1765--1778, 2016.

\bibitem[Tretschk et~al.(2021)Tretschk, Tewari, Golyanik, Zollh\"{o}fer,
  Lassner, and Theobalt]{tretschk2021nonrigid}
Edgar Tretschk, Ayush Tewari, Vladislav Golyanik, Michael Zollh\"{o}fer,
  Christoph Lassner, and Christian Theobalt.
\newblock Non-rigid neural radiance fields: Reconstruction and novel view
  synthesis of a dynamic scene from monocular video.
\newblock In \emph{{IEEE} International Conference on Computer Vision
  ({ICCV})}. {IEEE}, 2021.

\bibitem[Wang et~al.(2021)Wang, Geiger, and Tang]{Wang2021CVPR}
Shaofei Wang, Andreas Geiger, and Siyu Tang.
\newblock Locally aware piecewise transformation fields for 3d human mesh
  registration.
\newblock In \emph{Conference on Computer Vision and Pattern Recognition
  (CVPR)}, 2021.

\bibitem[Wang et~al.(2022)Wang, Schwarz, Geiger, and Tang]{ARAH:ECCV:2022}
Shaofei Wang, Katja Schwarz, Andreas Geiger, and Siyu Tang.
\newblock Arah: Animatable volume rendering of articulated human sdfs.
\newblock In \emph{European Conference on Computer Vision}, 2022.

\bibitem[Wang et~al.(2023)Wang, Han, Habermann, Daniilidis, Theobalt, and
  Liu]{wang2023neus2}
Yiming Wang, Qin Han, Marc Habermann, Kostas Daniilidis, Christian Theobalt,
  and Lingjie Liu.
\newblock Neus2: Fast learning of neural implicit surfaces for multi-view
  reconstruction.
\newblock In \emph{Proceedings of the IEEE/CVF International Conference on
  Computer Vision}, pages 3295--3306, 2023.

\bibitem[Weng et~al.(2022)Weng, Curless, Srinivasan, Barron, and
  Kemelmacher-Shlizerman]{weng_humannerf_2022_cvpr}
Chung-Yi Weng, Brian Curless, Pratul~P. Srinivasan, Jonathan~T. Barron, and Ira
  Kemelmacher-Shlizerman.
\newblock Human{N}e{RF}: Free-viewpoint rendering of moving people from
  monocular video.
\newblock In \emph{Proceedings of the IEEE/CVF Conference on Computer Vision
  and Pattern Recognition (CVPR)}, pages 16210--16220, 2022.

\bibitem[Wu et~al.(2023)Wu, Yi, Fang, Xie, Zhang, Wei, Liu, Tian, and
  Wang]{wu20234d}
Guanjun Wu, Taoran Yi, Jiemin Fang, Lingxi Xie, Xiaopeng Zhang, Wei Wei, Wenyu
  Liu, Qi Tian, and Xinggang Wang.
\newblock 4d gaussian splatting for real-time dynamic scene rendering.
\newblock \emph{arXiv preprint arXiv:2310.08528}, 2023.

\bibitem[Xu et~al.(2019{\natexlab{a}})Xu, Cheng, Guo, Han, Liu, and
  Fang]{FlyFusion}
Lan Xu, Wei Cheng, Kaiwen Guo, Lei Han, Yebin Liu, and Lu Fang.
\newblock Flyfusion: Realtime dynamic scene reconstruction using a flying depth
  camera.
\newblock \emph{IEEE transactions on visualization and computer graphics},
  27\penalty0 (1):\penalty0 68--82, 2019{\natexlab{a}}.

\bibitem[Xu et~al.(2019{\natexlab{b}})Xu, Su, Han, Yu, Liu, and
  Fang]{UnstructureLan}
Lan Xu, Zhuo Su, Lei Han, Tao Yu, Yebin Liu, and Lu Fang.
\newblock Unstructuredfusion: realtime 4d geometry and texture reconstruction
  using commercial rgbd cameras.
\newblock \emph{IEEE transactions on pattern analysis and machine
  intelligence}, 42\penalty0 (10):\penalty0 2508--2522, 2019{\natexlab{b}}.

\bibitem[Yang et~al.(2023{\natexlab{a}})Yang, Gao, Zhou, Jiao, Zhang, and
  Jin]{yang2023deformable}
Ziyi Yang, Xinyu Gao, Wen Zhou, Shaohui Jiao, Yuqing Zhang, and Xiaogang Jin.
\newblock Deformable 3d gaussians for high-fidelity monocular dynamic scene
  reconstruction.
\newblock \emph{arXiv preprint arXiv:2309.13101}, 2023{\natexlab{a}}.

\bibitem[Yang et~al.(2023{\natexlab{b}})Yang, Yang, Pan, Zhu, and
  Zhang]{yang2023real}
Zeyu Yang, Hongye Yang, Zijie Pan, Xiatian Zhu, and Li Zhang.
\newblock Real-time photorealistic dynamic scene representation and rendering
  with 4d gaussian splatting.
\newblock \emph{arXiv preprint arXiv:2310.10642}, 2023{\natexlab{b}}.

\bibitem[Yu et~al.(2019)Yu, Zheng, Guo, Zhao, Dai, Li, Pons-Moll, and
  Liu]{DoubleFusion}
Tao Yu, Zerong Zheng, Kaiwen Guo, Jianhui Zhao, Qionghai Dai, Hao Li, Gerard
  Pons-Moll, and Yebin Liu.
\newblock Doublefusion: Real-time capture of human performances with inner body
  shapes from a single depth sensor.
\newblock \emph{Transactions on Pattern Analysis and Machine Intelligence
  (TPAMI)}, 2019.

\bibitem[Yu et~al.(2021)Yu, Zheng, Guo, Liu, Dai, and Liu]{yu2021function4d}
Tao Yu, Zerong Zheng, Kaiwen Guo, Pengpeng Liu, Qionghai Dai, and Yebin Liu.
\newblock Function4d: Real-time human volumetric capture from very sparse
  consumer rgbd sensors.
\newblock In \emph{Proceedings of the IEEE/CVF Conference on Computer Vision
  and Pattern Recognition}, pages 5746--5756, 2021.

\bibitem[Zhang et~al.(2023)Zhang, Lin, Shao, Zhang, Zheng, Huang, Guo, and
  Liu]{zhang2023closet}
Hongwen Zhang, Siyou Lin, Ruizhi Shao, Yuxiang Zhang, Zerong Zheng, Han Huang,
  Yandong Guo, and Yebin Liu.
\newblock Closet: Modeling clothed humans on continuous surface with explicit
  template decomposition.
\newblock In \emph{Proceedings of the IEEE Conference on Computer Vision and
  Pattern Recognition}, 2023.

\bibitem[Zhao et~al.(2022{\natexlab{a}})Zhao, Jiang, Yao, Zhang, Wang, Dai,
  Zhong, Zhang, Wu, Xu, et~al.]{zhao2022human}
Fuqiang Zhao, Yuheng Jiang, Kaixin Yao, Jiakai Zhang, Liao Wang, Haizhao Dai,
  Yuhui Zhong, Yingliang Zhang, Minye Wu, Lan Xu, et~al.
\newblock Human performance modeling and rendering via neural animated mesh.
\newblock \emph{ACM Transactions on Graphics (TOG)}, 41\penalty0 (6):\penalty0
  1--17, 2022{\natexlab{a}}.

\bibitem[Zhao et~al.(2022{\natexlab{b}})Zhao, Yang, Zhang, Lin, Zhang, Yu, and
  Xu]{zhao2022humannerf}
Fuqiang Zhao, Wei Yang, Jiakai Zhang, Pei Lin, Yingliang Zhang, Jingyi Yu, and
  Lan Xu.
\newblock Humannerf: Efficiently generated human radiance field from sparse
  inputs.
\newblock In \emph{Proceedings of the IEEE/CVF Conference on Computer Vision
  and Pattern Recognition}, pages 7743--7753, 2022{\natexlab{b}}.

\bibitem[Zheng et~al.(2023)Zheng, Zhao, Zhang, Liu, and
  Liu]{zheng2023avatarrex}
Zerong Zheng, Xiaochen Zhao, Hongwen Zhang, Boning Liu, and Yebin Liu.
\newblock Avatarrex: Real-time expressive full-body avatars.
\newblock \emph{ACM Transactions on Graphics (TOG)}, 42\penalty0 (4), 2023.

\bibitem[Zollh{\"o}fer et~al.(2014)Zollh{\"o}fer, Nie{\ss}ner, Izadi, Rehmann,
  Zach, Fisher, Wu, Fitzgibbon, Loop, Theobalt, et~al.]{zollhofer2014real}
Michael Zollh{\"o}fer, Matthias Nie{\ss}ner, Shahram Izadi, Christoph Rehmann,
  Christopher Zach, Matthew Fisher, Chenglei Wu, Andrew Fitzgibbon, Charles
  Loop, Christian Theobalt, et~al.
\newblock {Real-time Non-rigid Reconstruction using an RGB-D Camera}.
\newblock \emph{ACM Transactions on Graphics (TOG)}, 33\penalty0 (4):\penalty0
  156, 2014.

\end{thebibliography}
}

\end{document}